\documentclass{opt2023}

\DeclareMathOperator*{\argmin}{{arg\,min}}
\def\params{w}
\def\RR{\mathbb{R}}
\usepackage{enumitem}

\title[Stepsize Tuning and Progressive Sharpening]{On the Interplay Between Stepsize Tuning\\ and Progressive Sharpening}

\optauthor{%
\Name{Vincent Roulet} \Email{vroulet@google.com}\\
\Name{Atish Agarwala} \Email{thetish@google.com}\\
\Name{Fabian Pedregosa} \Email{pedregosa@google.com}\\
\addr Google DeepMind}

\begin{document}

\maketitle

\begin{abstract}%

  Recent empirical work has revealed an intriguing property of deep learning
  models by which the sharpness (largest eigenvalue of the Hessian) increases
  throughout optimization until it stabilizes around a critical value at which
  the optimizer operates at the edge of stability, given a \emph{fixed}
  stepsize~\citep{cohen_gradient_2022}. We investigate empirically how the
  sharpness evolves when using stepsize-tuners, the Armijo linesearch and Polyak
  stepsizes, that adapt the stepsize along the iterations to local quantities
  such as, implicitly, the sharpness itself. We find that the surprisingly poor
  performance of a classical Armijo linesearch in the deterministic setting may
  be well explained by its tendency to ever-increase the sharpness of the
  objective. On the other hand, we observe that Polyak stepsizes operate
  generally at the edge of stability or even slightly beyond, outperforming its
  Armijo and constant stepsizes counterparts in the deterministic setting. We
  conclude with an analysis that suggests unlocking stepsize tuners requires an
  understanding of the joint dynamics of the step size and the sharpness.

\end{abstract}

\section{Introduction}

Selecting a good learning rate is a time-consuming part of practical machine learning.
To alleviate this, 
a number of automatic stepsize tuners have been proposed in the literature.
Two of the most common methods are the stochastic Armijo
line-search of 
\citet{vaswani2019painless} and the stochastic Polyak stepsize of 
\citet{berrada2020training, loizou2021stochastic}.
These methods enjoy strong theoretical guarantees that, as we will see, do not always 
translate into fast empirical performance. 

A different line of work has highlighted a number of important phenomena during training of deep neural networks with large
learning
rates~\citep{neyshabur_exploring_2017, gilmer_loss_2022, ghorbani_investigation_2019, foret_sharpnessaware_2022}.
In particular, many experiments have shown the tendency for the loss's sharpness (largest Hessian eigenvalue) to 
increase until it reaches the \emph{edge of stability} (EOS), in which the maximum 
eigenvalue increases to and stabilizes at $2/\gamma$ for stepsize $\gamma$,
corresponding to the largest eigenvalue that converges for gradient descent with
stepsize $\gamma$ in a quadratic objective~\citep{wu_how_2018, giladi_stability_2020,cohen_gradient_2022, cohen_adaptive_2022}.

\begin{figure}[ht]
    \centering
    \includegraphics[width=0.32\linewidth]{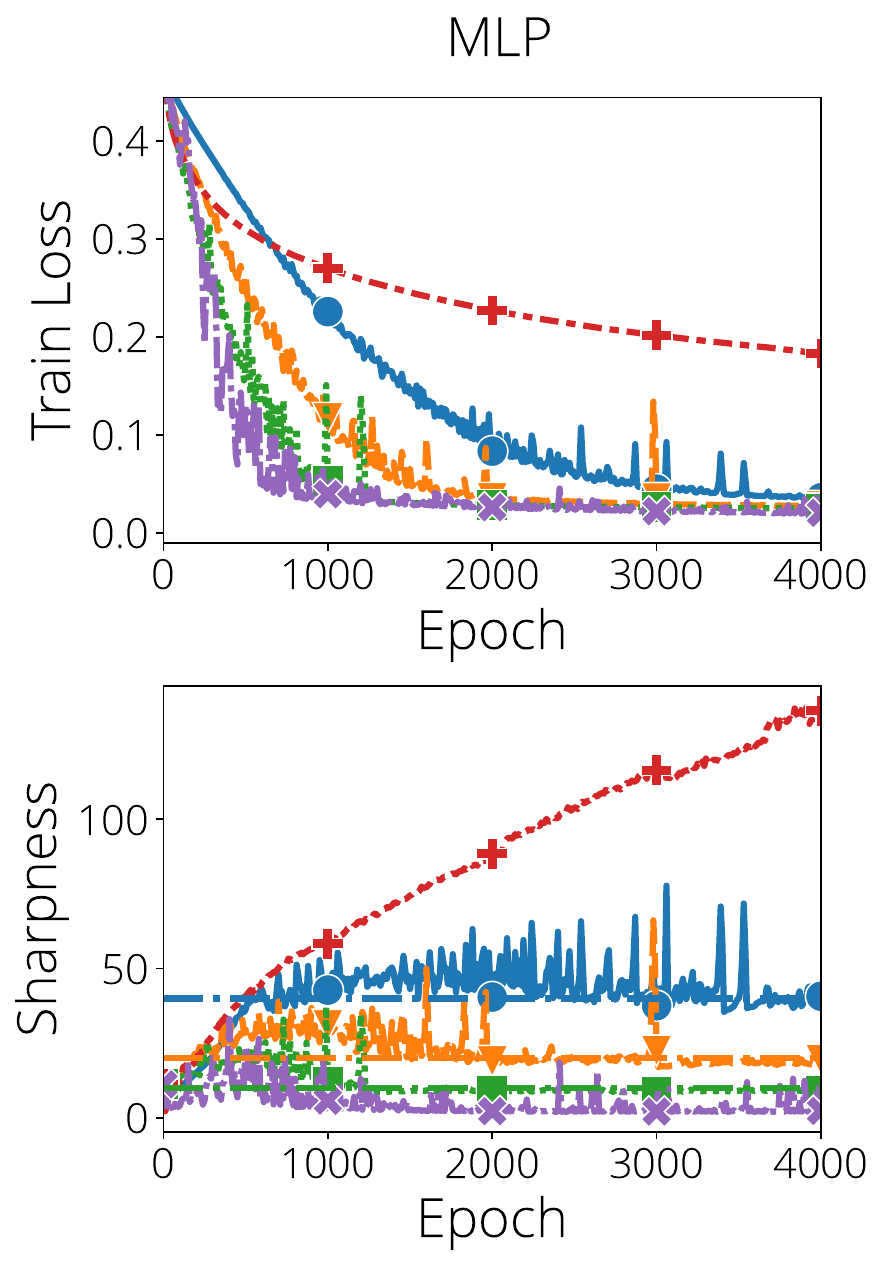}~
    \includegraphics[width=0.32\linewidth]{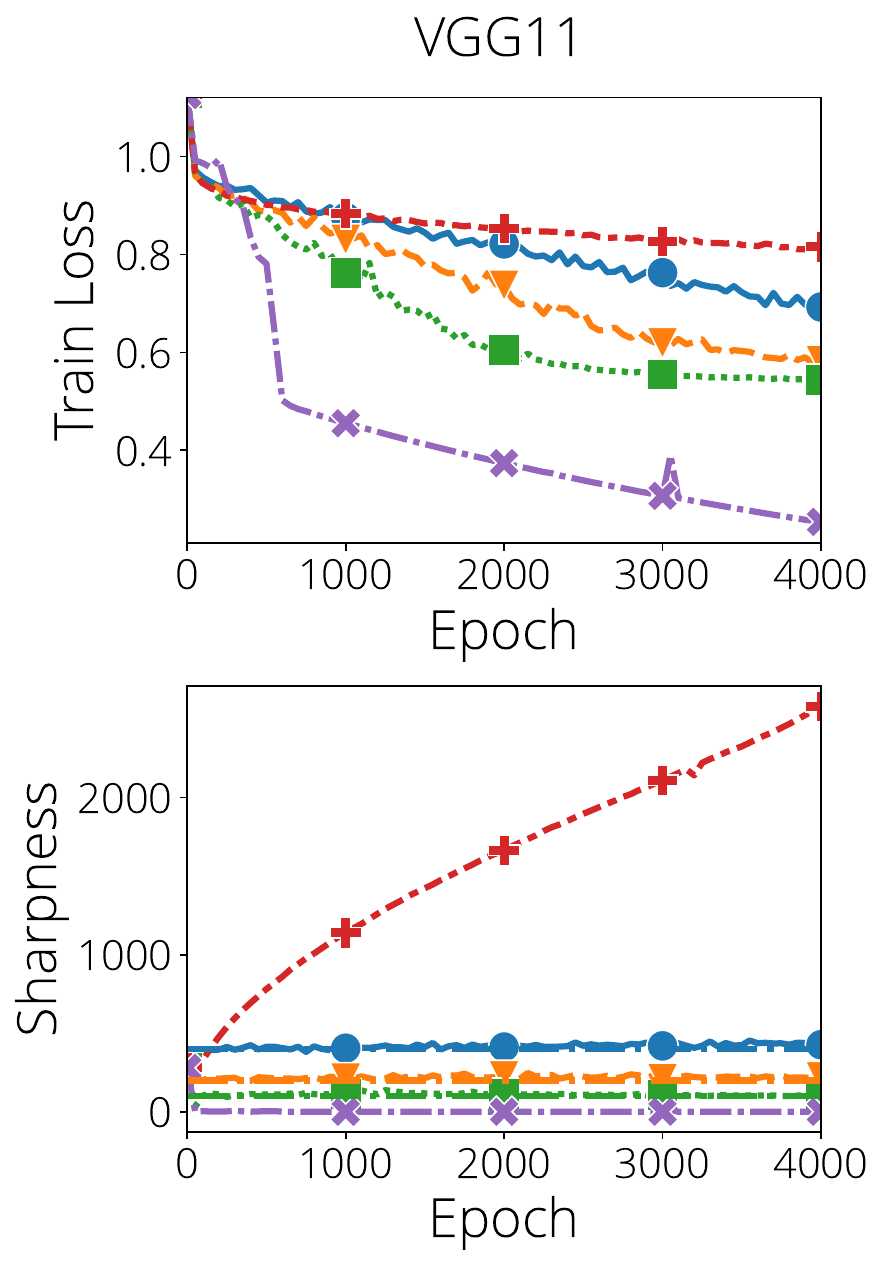}~
    \includegraphics[width=0.32\linewidth]{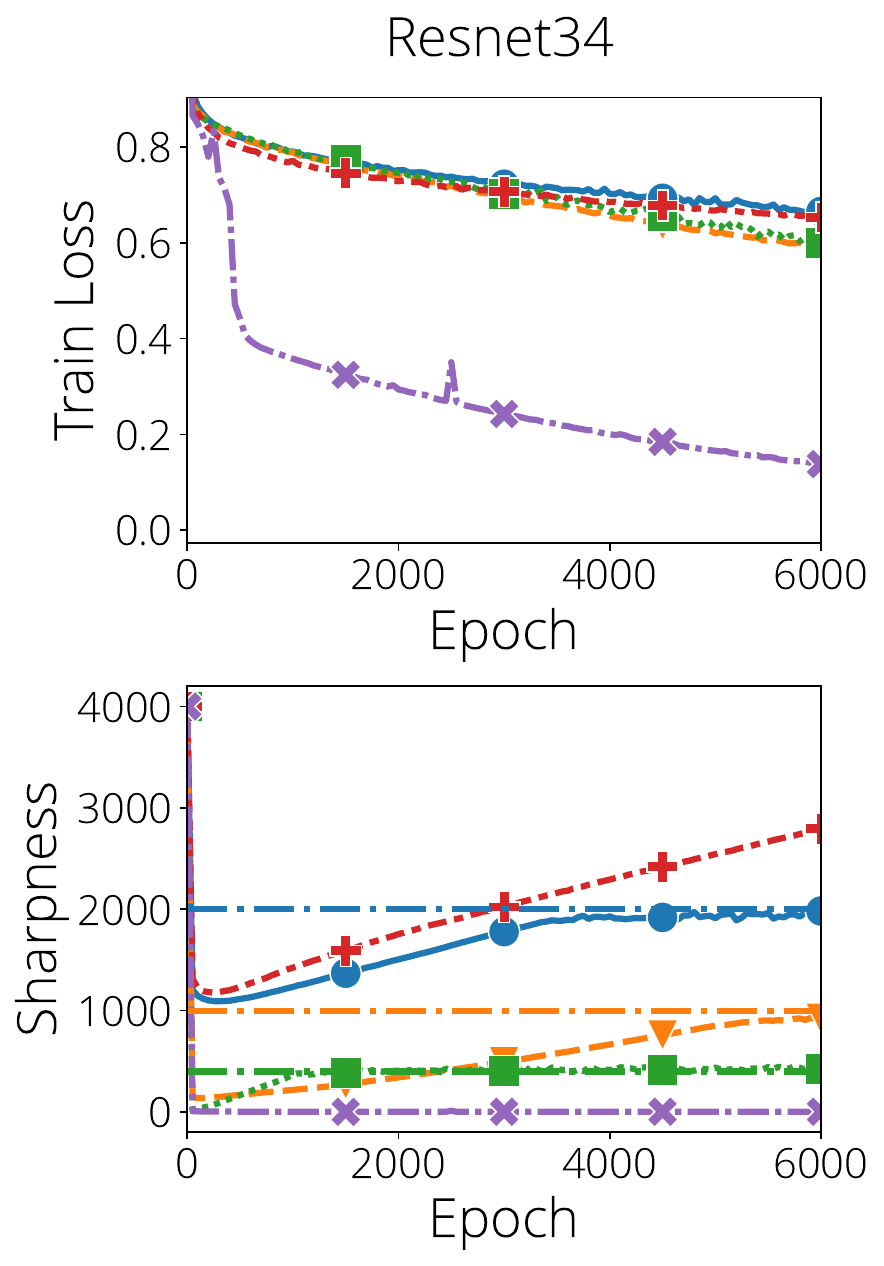}
    \includegraphics[width=\linewidth]{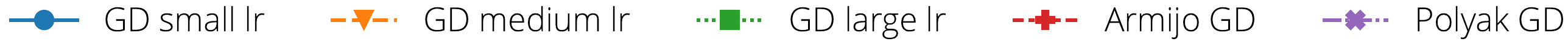}
    \caption{
{\bfseries Interplay between stepsize tuners and sharpness.} We plot the training loss (top) and sharpness (bottom) for various architectures on a subset of CIFAR10, for the different stepsize tuners (Armijo-GD, Polyak-GD) as well as GD with fixed stepsizes $\gamma$, all in the full-batch setting. The sharpness of GD stabilizes around the value $2/\gamma$ (dashed line). While Armijo-GD decreases the objective monotonically, the sharpness climbs further above any other method. In contrast, the train loss of Polyak-GD is not monotonically decreasing but the sharpness plateaus at low values.
}
    \label{fig:cifar10_full_batch_train_sharp}
\end{figure}

While there are some studies of sharpness dynamics with preconditioners
\citep{cohen_adaptive_2022}, little is known about sharpness dynamics with stepsize
tuners that condition on local properties of the loss landscape. This work takes a first
step to fill this void by providing a systematic study of loss and sharpness dynamics
in Armijo and Polyak stepsize tuners. 

\begin{minipage}{\linewidth}
We find the following:
\begin{itemize}[nosep]
\item In the \emph{deterministic} (full batch) setting, Armijo performs worse than constant stepsize while Polyak performs well. Armijo exhibits an ever-increasing sharpness while Polyak operates almost systematically at EOS or above.
\item In the \emph{stochastic} (minibatch) setting, the performance of Armijo depends highly on the mini-batch size, and can retrieve the performance analyzed by~\citet{vaswani2019painless}, while Polyak performs slightly less well than in the full batch setting and operates mostly again at EOS or above.
\item Our theoretical insights shows that the \emph{joint} dynamics of loss and stepsize
tuning process itself is necessary to explain these phenomena.
\end{itemize}
Our work suggests that design of stepsize tuning in deep learning must incorporate
sharpness dynamics.
\end{minipage}

\section{Methods}

We consider minimizing of the average $f$ of $n$ functions $f_1, \ldots, f_n$, representing typically the average loss of a deep network on a set of samples, that is, we aim to solve
\begin{equation}\label{eq:opt}
\params^{\star} \in \argmin_{\params \in \RR^p} \left\{ f(\params) = \frac{1}{n} \sum_{i=1}^n f_i(\params) \right\}\,.
\end{equation}
In that purpose, we consider gradient or stochastic gradient descent algorithms of the form
\begin{equation}
\params_{t+1} = \params_t - \gamma_t \nabla \phi(\params_t), \
\quad \mbox{with}\ \phi = f \ \mbox{(deterministic regime)} \ \mbox{or} \ \phi = f_i \ \mbox{(stochastic regime)}\,,
\end{equation}
for $i$ sampled uniformly at random in $[n]$ in the stochastic regime.
Here rather fixing the stepsize (a.k.a. learning rate) $\gamma_t$ to a constant, we consider stepsize-tuners that automatically adjust $\gamma_t$ given local information on the objective.
We consider two stepsize-tuners: Armijo line-search and Polyak stepsize.

\paragraph{Armijo line-search.} The Armijo line-search is a standard method for setting the stepsize of gradient-based optimizers in the deterministic setting \citep{nocedal1999numerical}. \citet{vaswani2019painless} adapted this classical method to the stochastic regime and provided convergence guarantees for smooth convex functions in the interpolation regime. 
The Armijo condition consists in selecting the stepsize $\gamma_t$ to ensure decrease of the objective or its mini-batch counterpart as
\begin{equation}\label{eq:armijo}
\phi(\params_t - \gamma_t \nabla \phi(\params_t)) 
\leq \phi(\params_t) - c \gamma_t \|\nabla \phi(\params_t)\|^2, 
\quad \mbox{with}\ \phi = f \ \mbox{(deterministic)} \ \mbox{or} \ \phi = f_i \ \mbox{(stochastic)}\,,
\end{equation}
for $c > 0$ a hyperparameter of the method, set to $10^{-4}$ in our experiments as in usual implementations. 
To ensure that the above is satisfied, one starts from a maximum stepsize $\gamma_{\max}$, set to $\gamma_{\max} =1$ in the experiments, and decreases the stepsize by a constant factor of $0.8$ until the above equation is satisfied. 

\paragraph{Polyak stepsize.} A recent breakthrough in optimization was the realization that the venerable Polyak stepsize, \citep{polyak1969minimization} originally developed for deterministic/full gradient optimization, extends naturally to stochastic optimization \citep{berrada2020training, loizou2021stochastic, jiang2023adaptive}. 
Many variants of this stepsize have been proposed, see, e.g.,~\citep{gower2022cutting, li2022sp2}. In this manuscript we focus on the $\text{SPS}_{\max}$ variant of \citep{loizou2021stochastic} that sets the stepsize based on the following formula:
\begin{equation}\label{eq:polyak}
    \gamma_t = \min\left\{\frac{\phi(w_t) - \phi^\star}{\|\nabla \phi(w_t)\|^2}, \gamma_{\max} \right\} \quad \mbox{with}\ \phi = f \ \mbox{(deterministic)} \ \mbox{or} \ \phi = f_i \ \mbox{(stochastic)}\,,
\end{equation}
for $\gamma_{\max}$ a hyperparameter of the method that restricts Polyak from taking a too large step. In our experiments, this parameter is set to $\gamma_{\max}=1$ and we consider $\phi^\star = 0$, that is, the model can overfit the training regime.

\section{Experiments and Discussion}

\begin{figure}[t]
    \centering
    \includegraphics[width=0.32\linewidth]{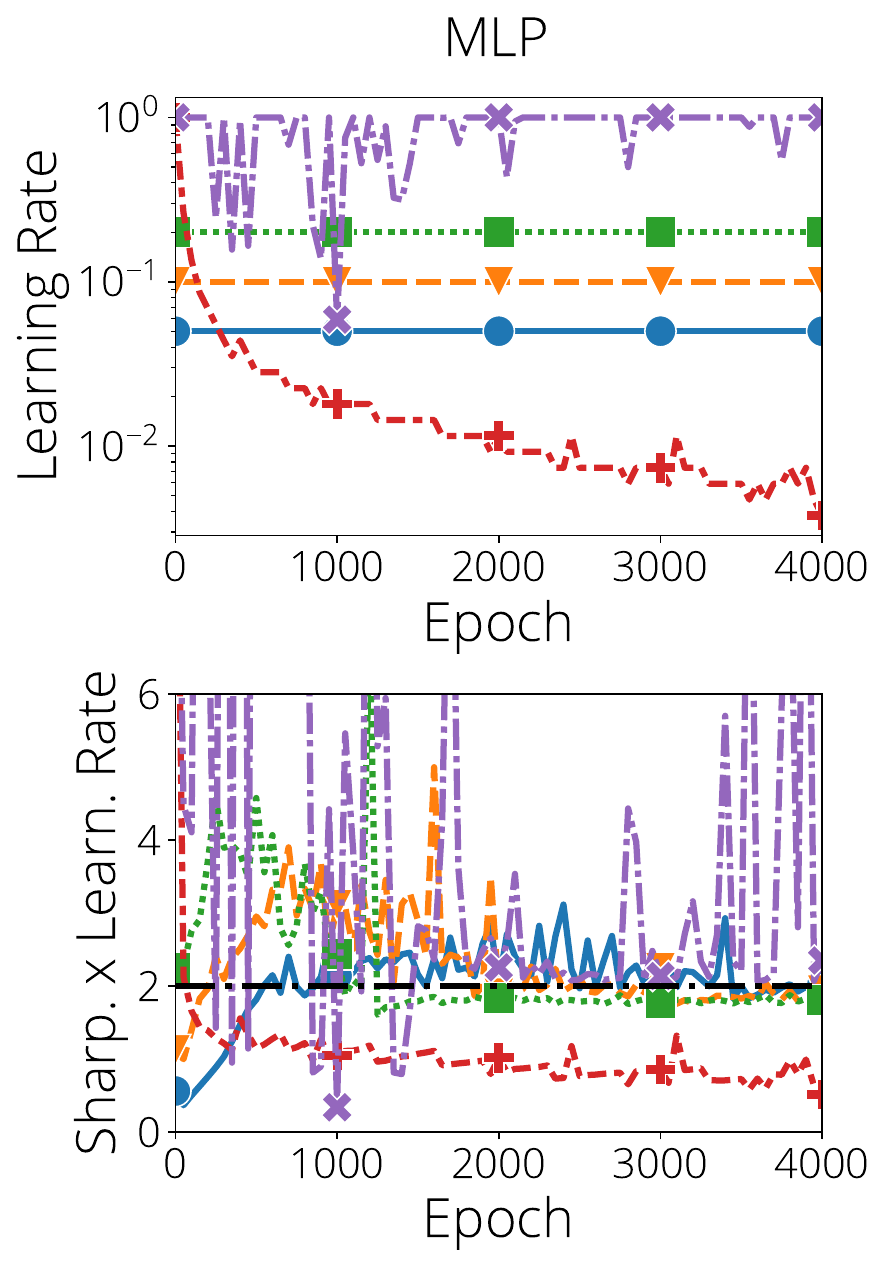}~
    \includegraphics[width=0.32\linewidth]{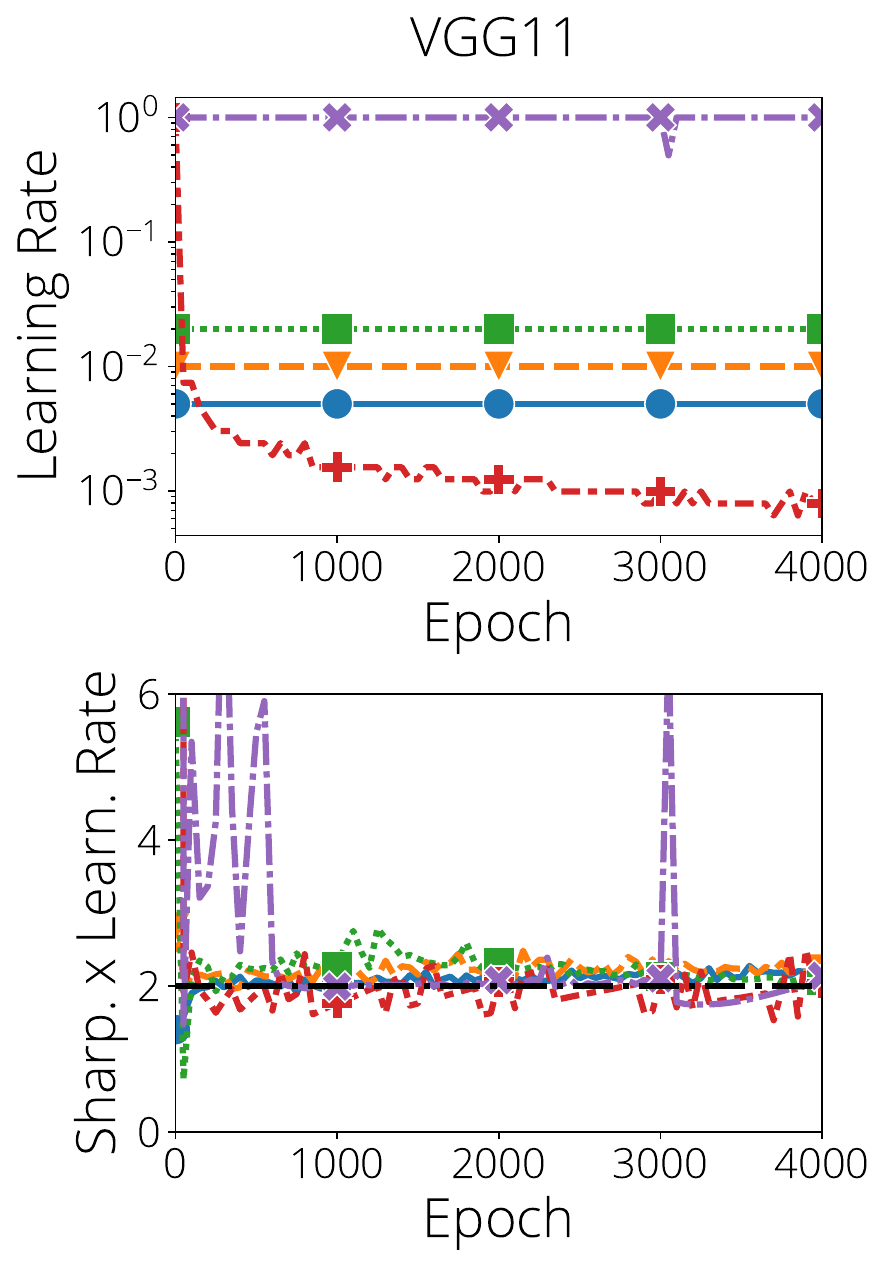}~
    \includegraphics[width=0.32\linewidth]{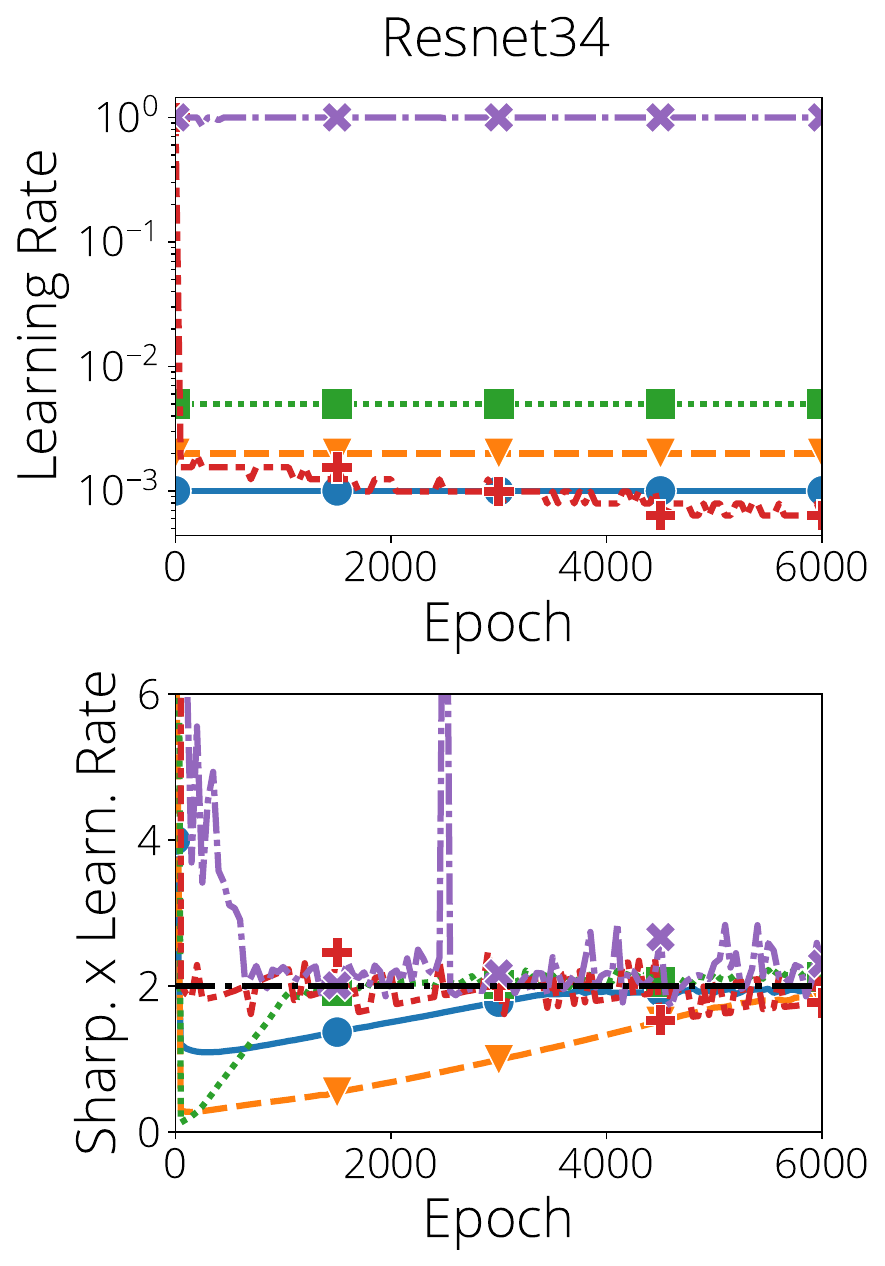}
    \includegraphics[width=\linewidth]{images/cifar_full_batch_train_sharp_legend.pdf}
    \caption{
    {\bfseries A closer look at the learning rate dynamics.} 
    In the top plot, we show the learning rate dynamics for the different stepsize tuners and GD. The Armijo backtracking line-search (red) decreases the learning rate monotonically, while the Polyak stepsize (violet) oscillates around the maximal acceptable value from~\eqref{eq:polyak}, $\gamma_{\max} = 1$. 
    In the bottom plot, we show the product of the learning rate and the sharpness of the Hessian. For constant stepsize GD, this product stabilizes at the critical EOS value $2$ (black dashed line). For Armijo-GD the value either stays well below $2$ or lingers around it from the start. For Polyak-GD, the product oscillates around the critical value without stabilizing like GD.    
    \label{fig:cifar_mlp_full_batch_lr_sharp}
    }
\end{figure}

We study stepsize tuners on the CIFAR10 image classification dataset~\citep{krizhevsky2009learning} with a squared loss and weight decay
with parameter $10^{-4}$. We will distinguish between the \emph{deterministic}
setting -- full batch gradient descent (GD) -- and the \emph{stochastic}
setting -- minibatch gradient descent (SGD). We selected the stepsizes for GD to be close to the maximal stepsize that does not lead to divergence. 

\subsection{Deterministic regime}

\paragraph{Setting.}
We first consider training deep networks on a reduced subset of $4096$ samples of CIFAR10
training on all the samples each step. We trained on a $6$-layer fully connected
network, VGG, and with ResNet. Architectural details can be found in
Appendix \ref{app:architectures}.

\paragraph{Results.}
As promised, Armijo-GD (linesearch, deterministic setting) gives
monotonically decreasing loss for all
architectures (Fig.~\ref{fig:cifar10_full_batch_train_sharp}, top). This is in contrast
with the extremely non-monotonic loss trajectories for fixed stepsize gradient descent
(GD). Note that this non-monotonicity comes from non-linear discrete effects, and not
sampling noise. However, Armijo-GD greatly underperforms constant stepsize GD.
This observation comes somewhat as a surprise given that stepsize tuners such as Armijo 
have been the workhorse of non-convex optimization for 
decades~\citep{nocedal1999numerical}. This inefficiency seems to be related to the
strong progressive sharpness (increasing sharpness) of Armijo-GD
(Fig.~\ref{fig:cifar10_full_batch_train_sharp}, bottom). The cost of the non-monotonicity
in the face of increasing sharpness is a decrease in stepsize
(Fig~\ref{fig:cifar_mlp_full_batch_lr_sharp}, top). This can be further elucidated by
measuring the stepsize times the sharpness (Fig.~\ref{fig:cifar10_full_batch_train_sharp}, 
bottom). For fixed stepsize this oscillated above and below the EOS at value $2$, leading
to sharpness regularization; for Armijo-GD, the value is very much below $2$ for the MLP
(left, bottom), and near but often below $2$ for the other architectures.

In contrast, Polyak-GD gives non-monotonic loss trajectories but with a faster decrease than Armijo,
and leads to low final values of the sharpness (Fig~\ref{fig:cifar_mlp_full_batch_lr_sharp}).
The stepsizes selected by Polyak's method tend to select the the maximal 
acceptable value from~\eqref{eq:polyak}, $\gamma_{\max} = 1$, and leads
the normalized
spectral norm to oscillate about (and sometimes well above) the critical value of $2$ (Fig~\ref{fig:cifar_mlp_full_batch_lr_sharp}). The plot on Fig.~\ref{fig:cifar_mlp_full_batch_lr_sharp} registered the metrics every 50 epochs. Zooming in the first iterations of the method and recording at each iteration as done in Fig.~\ref{fig:zoom_polyak} in Appendix~\ref{app:more_exp} shows that Polyak's method only selects large stepsizes (around the maximal one) once the sharpness has sufficiently decreased.

\subsection{Stochastic regime}

\begin{figure}
    \centering
    \includegraphics[width=\linewidth]{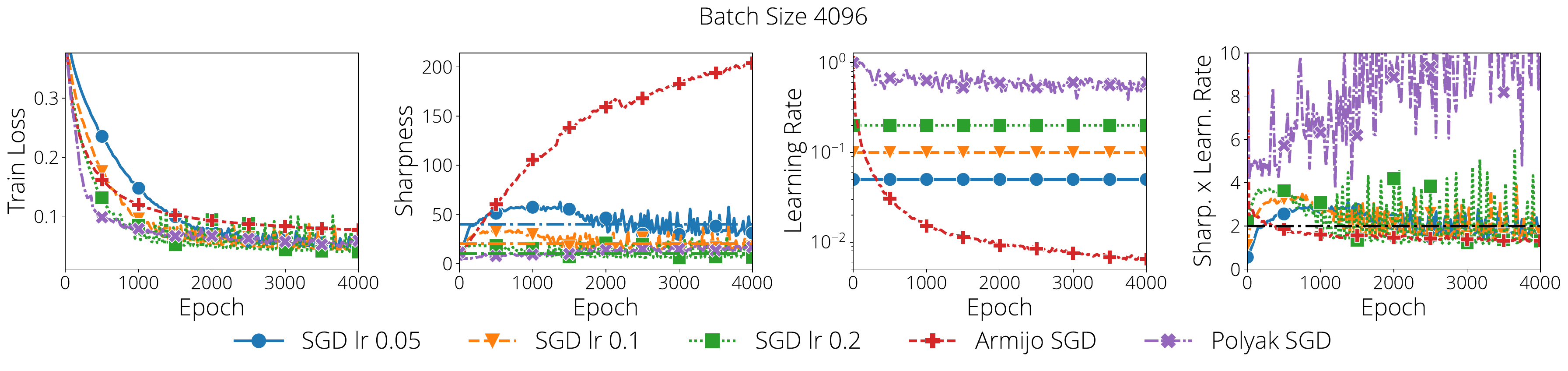}
    \includegraphics[width=\linewidth]{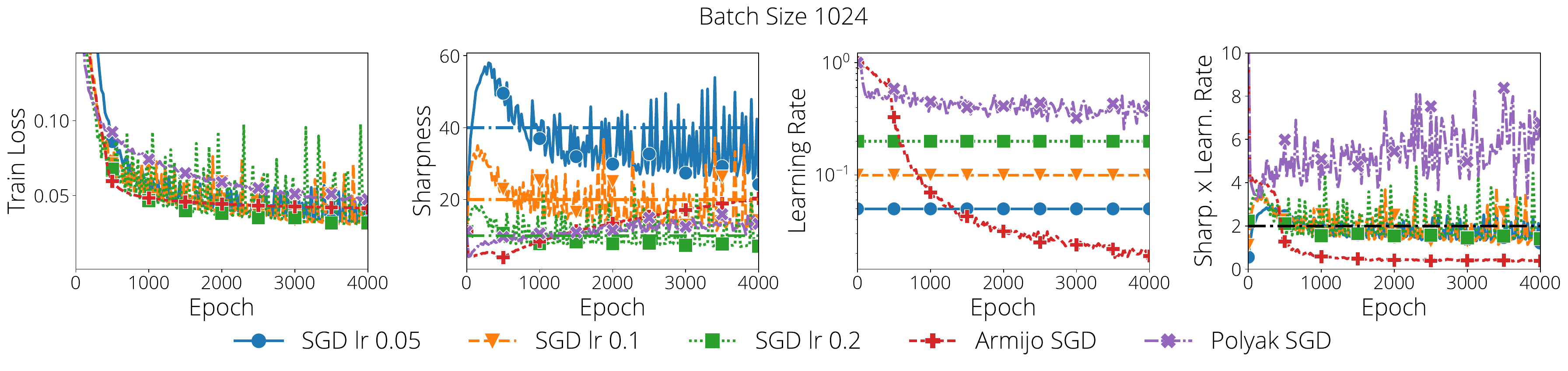}
    \includegraphics[width=\linewidth]{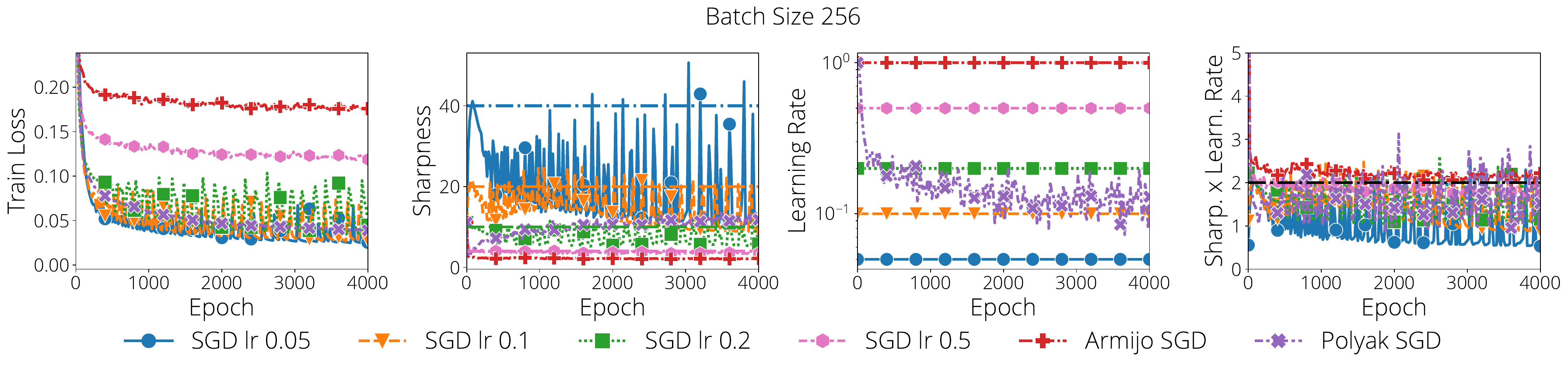}
    \caption{{\bfseries Batch sizes impact the behavior of stepsize tuners in the stochastic regime.}
    We plot training loss, sharpness, learning rate, and normalized sharpness for training an MLP (see Appendix for VGG and ResNet) on the full CIFAR10 dataset with various batch-sizes. In this stochastic regime, the performance of Armijo vary greatly with the mini-batch considered, with a good performance at medium scale, and poor performance otherwise. Progressive sharpening of Armijo is only observed at medium and large scales. At small scale, Armijo displays an ``instantaneous'' sharpening that hinders its performance just as with large stepsizes SGD. Polyak performs reasonably well in all settings, while operating potentially above the edge of stability regime in, e.g., the medium scale.
    }
    \label{fig:mlp_stochastic}
\end{figure}

\paragraph{Setting.}
We repeat the experiments, this time in the stochastic (minibatch) setting. We now
train on all of CIFAR10, with minibatch sizes $\{256, 1024, 4096\}$. Results in Fig.~\ref{fig:mlp_stochastic}
are in the MLP architecture while results on VGG and ResNet are presented in Appendix~\ref{app:more_exp}.

\paragraph{Results.} The performance of the stepsize tuners varies greatly with the
batch size. For the largest batch size of 4096 (Fig.~\ref{fig:mlp_stochastic}, top),
we observe essentially the same behavior as in the deterministic (full batch) setting:
Armijo-SGD displays progressive sharpening to long times, the stepsize decreases,
and optimization is poor. Polyak once again performs well.

For the intermediate batch size of 1024, (Fig.~\ref{fig:mlp_stochastic},
middle), Armijo-SGD optimizes well while maintaining relative smoothness of the
loss trajectory. There is sharpening but less than some of the GD
settings. Armijo-SGD still stays below the EOS. In contrast, Polyak-SGD now
optimizes similarly to Armijo and fixed step size SGD (slower at the start but
faster later) to similar values of the loss. Polyak-SGD seems to stabilize above
the EOS.

For the smallest batch size of $256$ (Fig.~\ref{fig:mlp_stochastic}, bottom), Armijo-SGD
induces small sharpness, but dramatically underperforms fixed stepsize SGD. In this case,
Armijo-SGD often selects the maximal stepsize\footnote{The plot is in epochs, at the scale of each steps, we observed stepsizes sometimes of the order of 0.5 rather than 1.}
and remains above EOS. This is in contrast to fixed stepsize SGD which
remains below the EOS. This is similar to Polyak-SGD which performs poorly here compared to fixed stepsize 
while staying above the EOS.

The varying behavior of Armijo-SGD in the stochastic regime may be explained by the discrepancy between the maximal acceptable stepsize on a mini-batch and on the full loss. As shown by \citet[Figure 7]{mutschler2020parabolic}, the best stepsize decreasing the value on a mini-batch tends to be larger than the stepsize that would have been selected on the full loss. The improved performance of Armijo in the stochastic setting may then be explained by its inability to satisfy the criterion that would be used in the full-batch regime.

\section{Simple models fail to capture stepsize tuner dynamics}\label{sec:theory}

In their empirical study, \citet{cohen_gradient_2022} uncovered two distinct phenomena:
increasing largest Hessian eigenvalue at early times (progressive sharpening) and
stabilization of said eigenvalue around a critical value (EOS). One natural hypothesis
is that good stepsize tuners will quickly reach values near the EOS and stay there
during training. We will analyze the stepsize tuners using a simple model of EOS
behavior and see that EOS alone is not sufficient to understand our experiments -
we must eventually understand the sharpening dynamics themselves.

The edge of stability dynamics is well represented by the
projection onto the top eigenmode of the Hessian/NTK jointly with the dynamics of the 
eigenvalue itself \citep{damian_selfstabilization_2022, agarwala_second_2023}. We ask a 
basic question: can this
simplified model be used to derive the stable values for the step sizes achieved by
Armijo-GD and Polyak-GD observed in practice?

More concretely, suppose $\params_t$ is close to a minimum $\params^{\star}$. Suppose that
$\params_t-\params^\star$ is given by $a_{t} v_{\max, t}$, where $a_{t}$ is a scalar and
$v_{\max, t}$ is the eigenvector associated with the largest eigenvalue $\lambda_{\max, t}$ of
$\nabla^{2}f(\params_t)$. The edge of stability dynamics corresponds to $\lambda_{\max, t}$
staying near $2/\gamma_t$ and $a_{t}$ oscillating about $0$. Under these assumptions,
the question is: what is the behavior of the constraints on $\gamma_{t}$ induced by
the stepsize tuners?

We will make the further approximation that the loss is well approximated by the second 
order expansion around $\params^\star$, and that $v_{t}$ and $\lambda_{\max, t}$ are
approximately constant. The loss and gradients are then approximated as
\begin{equation}
f(\params_t) \approx \frac{1}{2}(\params_t-\params^\star)^\top \nabla^{2}f(\params^\star) (\params_t-\params^\star) 
= \frac{1}{2}\lambda_{\max}a_{t}^2,
\quad
\nabla f(\params_t)\approx \lambda_{\max}a_{t} v_{\max}. \nonumber
\end{equation}
The Armijo-GD condition \eqref{eq:armijo} becomes:
\begin{equation}\nonumber
\frac{1}{2}\lambda_{\max}(1-\lambda_{\max}\gamma_t)^2a_{t}^2 \leq  \lambda_{\max}(1-c\gamma_t \lambda_{\max})a_{t}^2,
\end{equation}
which constrains the stepsize selected by Armijo to satisfy
\begin{equation}\nonumber
\gamma_t \leq \frac{2(1-c)}{\lambda_{\max}}.
\end{equation}
For slowly changing $\lambda_{\max, t} \approx \lambda_{\max}$, Armijo-SGD
stabilizes below the EOS. We observe in
Fig.~\ref{fig:armijo_continuous_decrease} in Appendix~\ref{app:more_exp} that by
selecting $c=0$ and a decreasing factor close to 1, the Armijo linesearch
performs much better. This suggests that targeting the edge of stability may
benefit the optimization.

For Polyak-GD we have, for $\gamma_{\max} = 1$,
\begin{equation}\nonumber
\gamma_{t} 
= \min\left\{\frac{f(\params_t) - f^\star}{\|\nabla f(\params_t)\|^2}, 1\right\} 
= \min\left\{\frac{1}{2\lambda_{\max}}, 1\right\} 
= \begin{cases}
    1/(2\lambda_{\max}) & \mbox{if}\  \lambda_{\max} < 1/2 \\
    1 & \mbox{otherwise} 
\end{cases}.
\end{equation}
This threshold is also below the EOS, and in fact smaller
than the threshold for Armijo-GD.

However, in practice Polyak-GD seems to give rise to
larger learning rates than Armijo-GD, while in the SGD setting things are highly senstitive to batch
size. Therefore this simple model does not capture the true dynamics. Understanding
stepsize tuners requires analysis of the \emph{joint} dynamics of
$\gamma_{t}$ and $\lambda_{\max, t}$; in other words, progressive sharpening must be
considered as well as EOS.

\section{Conclusions}

We investigated how the sharpness of a network evolves as it is trained by gradient
or stochastic gradient methods whose stepsizes adapt themselves to local properties 
of the network such as the sharpness. We first observed that a classical Armijo linesearch
in the deterministic regime fails to capture the best performance of its fixed stepsize 
counterpart, contrarily to its underlying rationale. By monitoring the sharpness, we provide
a reasonable cause for this phenomenon: the Armijo linsearch never reaches the EOS, keeps 
increasing the sharpness, while taking ever decreasing stepsizes. In stark contrast, Polyak 
stepsizes outperform qualitatively the constant stepsizes counterparts while remaining 
constantly at the EOS or slightly above.

A further investigation in the stochastic regime revealed that stepsize tuners may highly 
depend on the batch-sizes considered relegating the ``tuning'' of the algorithm to the actual 
batch-size considered. This phenomenon is particularly true for Armijo that can still display 
an ever-increasing sharpness for large batch-sizes. On the other hand, Polyak generally performs 
reasonably well, operating around the EOS regime, although its performance appears impacted 
by the stochasticity compared to the deterministic regime.

Our empirical results combined with our theoretical analysis suggest that in order to
improve stepsize tuners, methods must incorporate the existence of progressive sharpening
and edge of stability phenomena. These seem to couple with the dynamics of the stepsize
tuners themselves in a non-trivial way, and understanding this coupling may be the key
to unlocking the potential of stepsize tuners in deep learning.\\

\paragraph{Acknowledgements.}
The authors thank Peter Bartlett, Aleksandar Botev, Mathieu
Blondel, Philip Long, James Martens, Courtney Paquette, Jeffrey Pennington, for
their insights on the manuscript.

\clearpage
\bibliography{eos_tuner_refs}

\clearpage

\appendix
\section{Experimental details}
\subsection{Architectures}

\label{app:architectures}

Here are some additional details on the architectures considered
\begin{enumerate}
    \item \emph{MLP}: We considered 6 layers of output size $(128, 128, 64, 64, 32, 32)$ followed by a last classification layer of output dimension $10$. All layers are without batch normalization or dropout. Each layer consists in a linear transformation followed by the ReLU activation function except for the classification layer, which has no activation. 
    \item \emph{VGG11}: We follow the implementation of the VGG11 architecture~\citep{simonyan2014very} except that we removed dropout layers and the batch normalization layers.
    \item \emph{ResNet34}: We follow the implementation of the Resnet34 architecture~\citep{he2016deep} except that we removed the batch normalization layers.
\end{enumerate}

\subsection{Evaluation}
\begin{enumerate}
    \item \emph{Sharpness computation}: To compute the sharpness, that is, the spectral norm of the Hessian, we used a power iteration method, stopped once changes in the residual go below a threshold of $10^{-3}$ or after $1000$ iterations. 
    \item \emph{Visualization}: For the stochastic experiments we average the evaluations of each quantity around 20 steps.
\end{enumerate}

\section{Additional experiments} \label{app:more_exp}
\subsection{VGG and ResNet in stochastic regime}
\begin{figure}
    \centering
    \includegraphics[width=\linewidth]{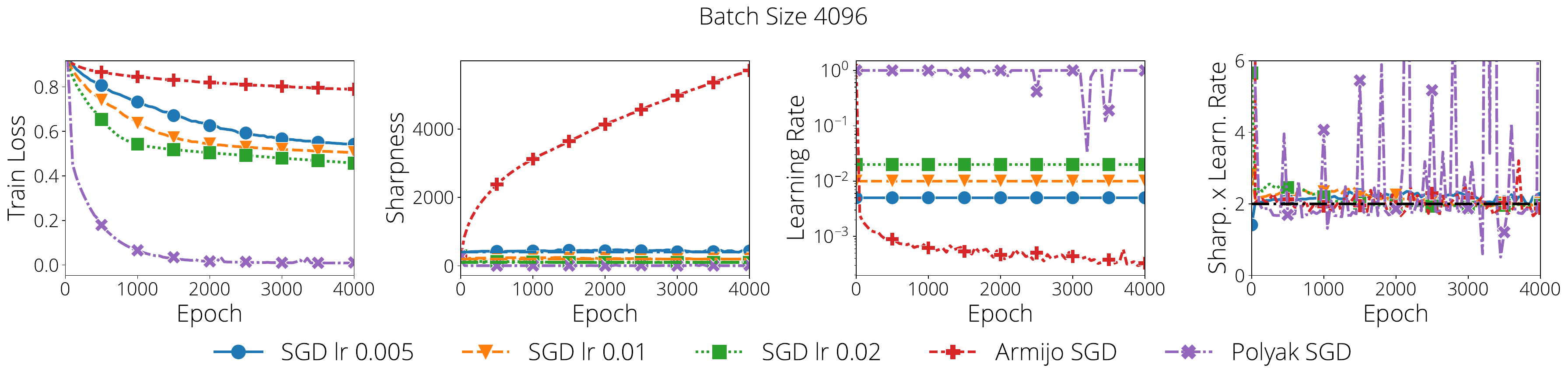}
    \includegraphics[width=\linewidth]{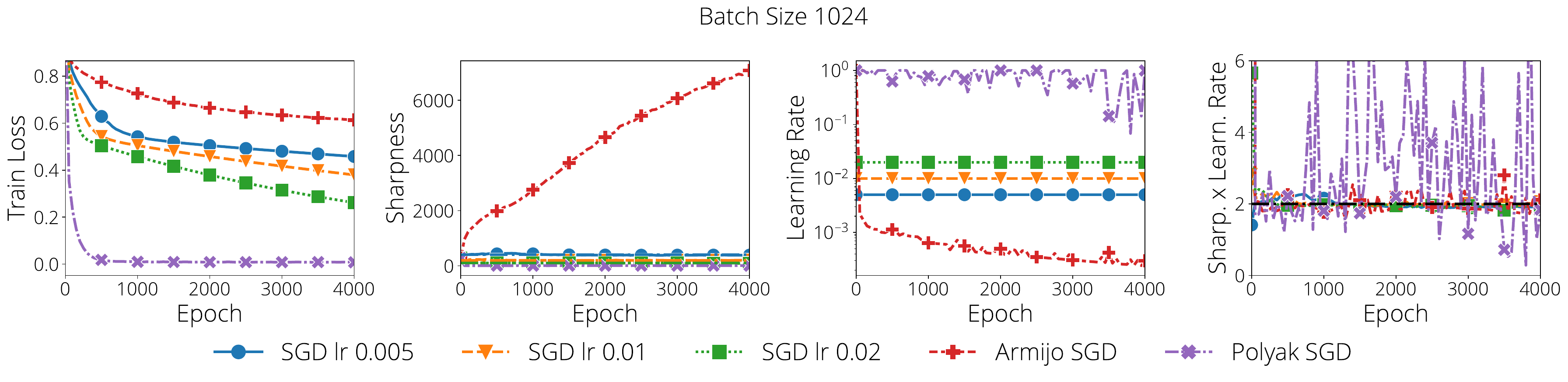}
    \includegraphics[width=\linewidth]{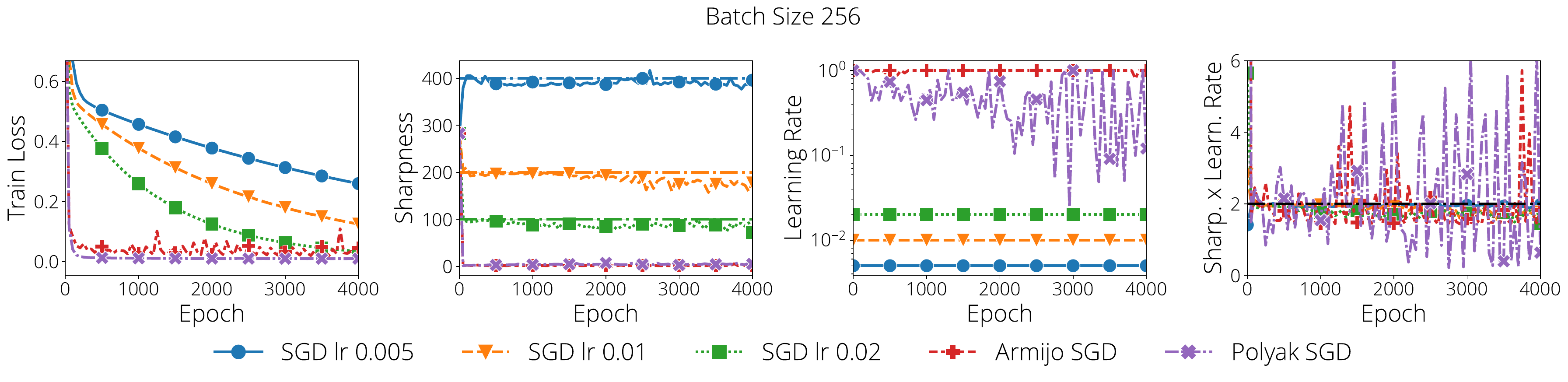}
    \caption{Training dynamics in stochastic regime for the VGG11 architecture.}
    \label{fig:cifar_vgg11_stoch_train_sharp}
\end{figure}

\begin{figure}
    \centering
    \includegraphics[width=\linewidth]{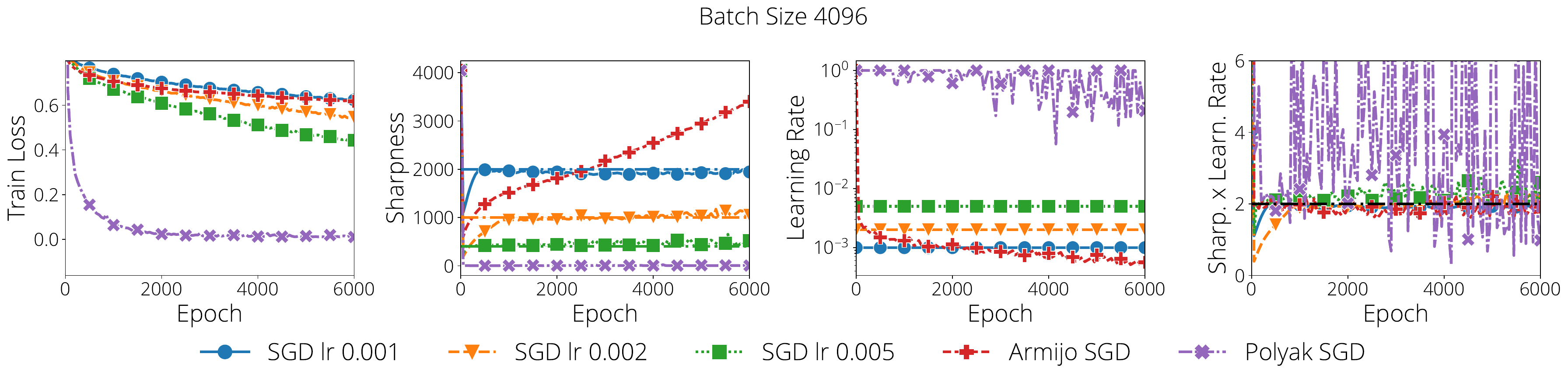}
    \includegraphics[width=\linewidth]{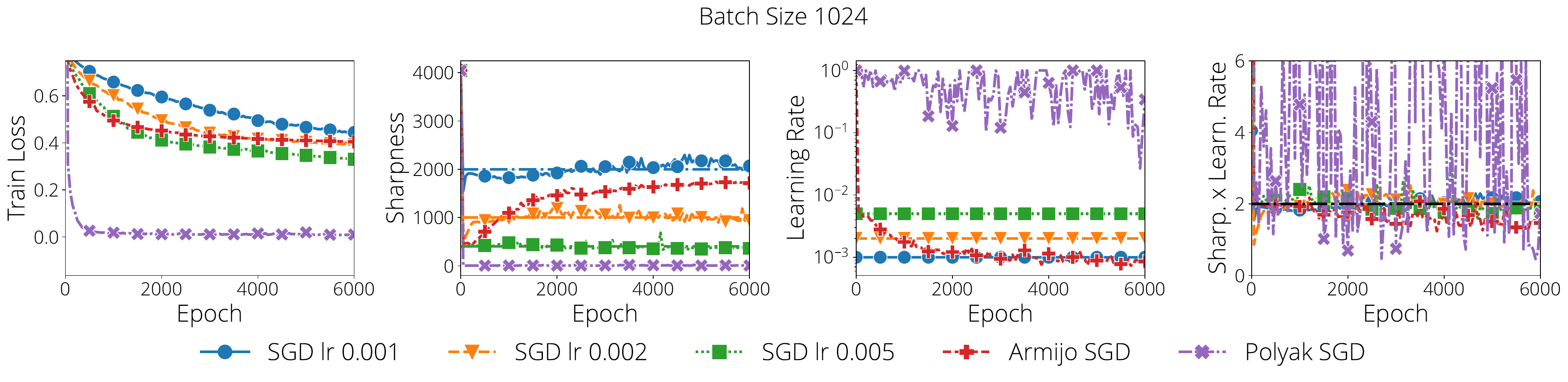}
    \includegraphics[width=\linewidth]{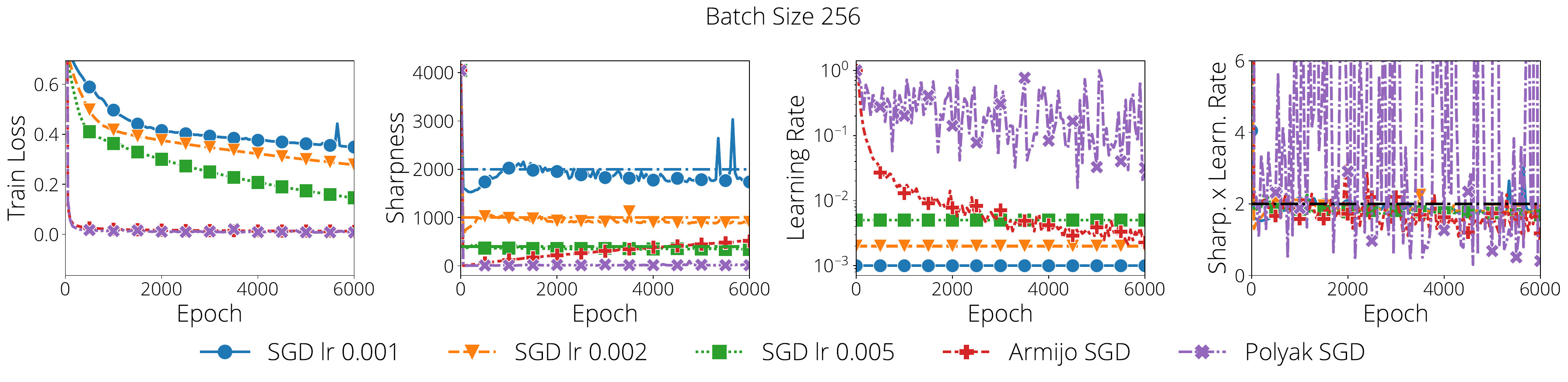}    
    \caption{Training dynamics in stochastic regime for the ResNet34 architecture.}
    \label{fig:cifar_resnet34_stoch_train_sharp}
\end{figure}

On Fig.~\ref{fig:cifar_vgg11_stoch_train_sharp} and~\ref{fig:cifar_resnet34_stoch_train_sharp}, we display the training dynamics of SGD, ArmijoSGD and PolyakSGD, when considering the VGG and ResNet architectures. Compared to the MLP setting of Fig~\ref{fig:mlp_stochastic}, we still observe the same sharpening Armijo at large and medium scales, though less starkly displayed for the ResNet architecture. In this regime, we oberve that the best regime for which Armijo works is the smallest one, hinting that the ideal batch-size for Armijo naturally depends on the architecture considered. On the other hand, Polyak SGD performs very well in all settings. Note that the constant stepsizes for SGD were chosen as in the full batch regime and do not illustrate the best possible performance of SGD in these settings. In other words, the stark contrast of performance between PolyakSGD and the constant size variants may be an effect of the loose search of constant stepsizes of SGD rather than a true qualitative difference. On the other hand, this illustrates the benefit of Polyak SGD in this context which alleviates the search for a stepsize. 

\subsection{Resnet34 with a logistic loss}
\begin{figure}
  \centering
  \includegraphics[width=\linewidth]{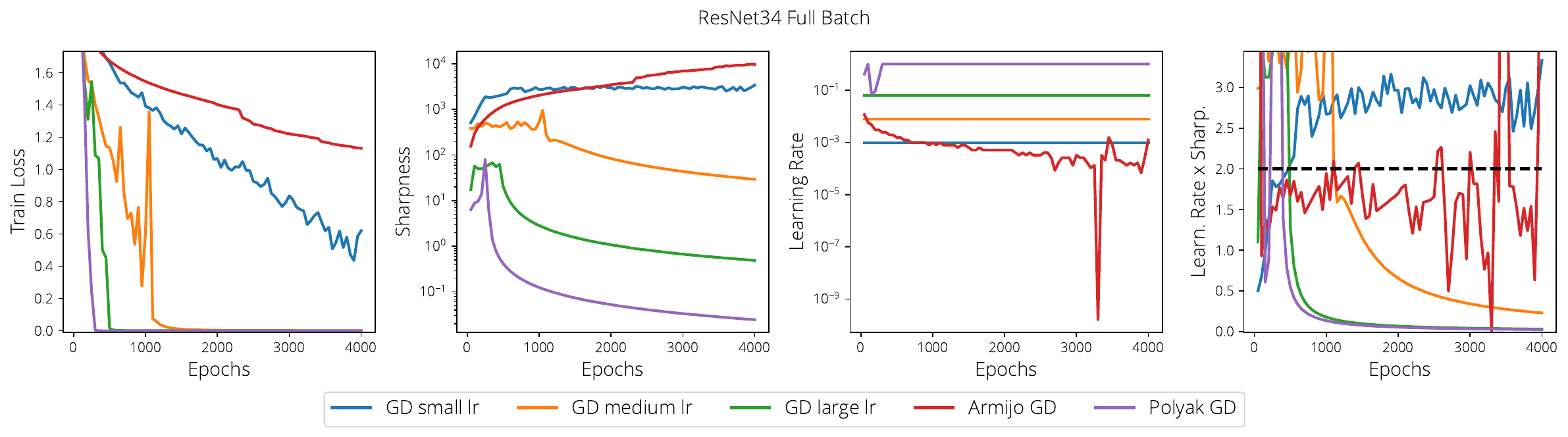}
  \caption{{\bfseries Armijo and Polyak behaves similarly with a logistic loss.}
  Training a ResNet34 on a subset of CIFAR10 (full batch) with a logistic loss.
  Armijo underperforms, Polyak outperforms constant learning rates GD. Sharpness
  vanishes in the long term except for Armijo and small learning rate GD.
  \label{fig:cifar10_resnet_logistic}}
\end{figure}

In Fig.~\ref{fig:cifar10_resnet_logistic}, we repeat the experiments of
Fig.~\ref{fig:cifar10_full_batch_train_sharp}
and~\ref{fig:cifar_mlp_full_batch_lr_sharp}, this time with a logistic loss
rather than the squared loss previously used. We focus on the ResNet34
architecture, on a subset of 4096 examples of CIFAR10, with no regularization
(we did not observe qualitative changes by changing the regularization).

We observe similar trends as for the squared loss: the Armijo underperforms
while Polyak outperforms the cosntant stepsizes GD (stepsizes were selected on a
logartihmic grid of base 2, and we selected three representative ones). The main
difference is that in this case, the sharpness tends to vanish rather than
converging to $2/\gamma$ for the constant stepsizes GD, a point already observed
by~\citet{cohen_gradient_2022}.

\subsection{Sensitivity to hyperparameters for Armijo}
We consider here varying some hyperparameters of the Armijo linesearch~\eqref{eq:armijo}.

\begin{figure}
  \centering
  \includegraphics[width=\linewidth]{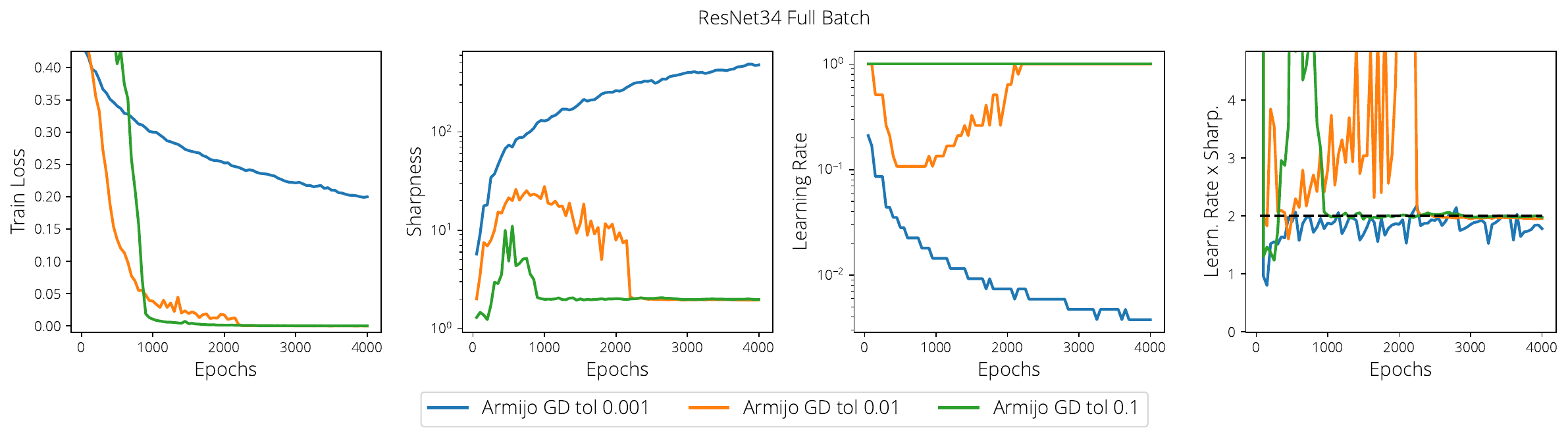}
  \caption{{\bfseries Loosening Armijo prevents its ever-increasing sharpness.}
  Armijo GD with a loose tolerance on a ResNet34,
  squared loss, subset of CIFAR10 (full batch). Loosening the Armijo condition
  enables Armijo to reach the EOS in the long term. \label{fig:armijo_loose}}
\end{figure}

\paragraph{Loose Armijo.}
We consider first relaxing the linesearch condition to accept stepsizes satisfying
\begin{equation}\label{eq:armijo_loose}
  \phi(\params_t - \gamma_t \nabla \phi(\params_t)) 
  \leq \phi(\params_t) - c \gamma_t \|\nabla \phi(\params_t)\|^2 + \varepsilon, 
  \quad \mbox{with}\ \phi = f \ \mbox{(deterministic)} \ \mbox{or} \ \phi = f_i \ \mbox{(stochastic)}\,,
\end{equation}
for $\varepsilon > 0$ some tolerance parameter.  
In Fig.~\ref{fig:armijo_loose}, we consider training a ResNet 34 on a subset of
CIFAR10 (full batch regime) with various values of the tolerance coefficient. We
observe that by loosening the tolerance, Armijo GD can aovid its ever-increasing
sharpness. However, we also found that selecting e.g. $\varepsilon=1$ led to
divergent behavior. One can already observe in~\ref{fig:armijo_loose} that a
loose tolerance of $\varepsilon=0.1$, is slower at the start and catches up the
$\varepsilon=0.01$ version only after some iterations. Note that in the long
term, both $\varepsilon =0.01$ and $\varepsilon=0.1$ reach the edge of stabiltiy
corresponding to taking $\gamma=1$.

\paragraph{Armijo with continuous decrease}
The analysis done in Section~\ref{sec:theory} suggests that Armijo could operate
at the edge of stability provided that we chose $c=0$. Another ingredient to
ensure that the stepsize is close to the edge of stability is to choose a
decreasing factor close to $1.$. In an extreme case, where we halve the stepsize
each time the condition is violated, the Armijo linesearch may miss the edge of
stability by a factor $1/2$.

\begin{figure}
  \centering
  \includegraphics[width=\linewidth]{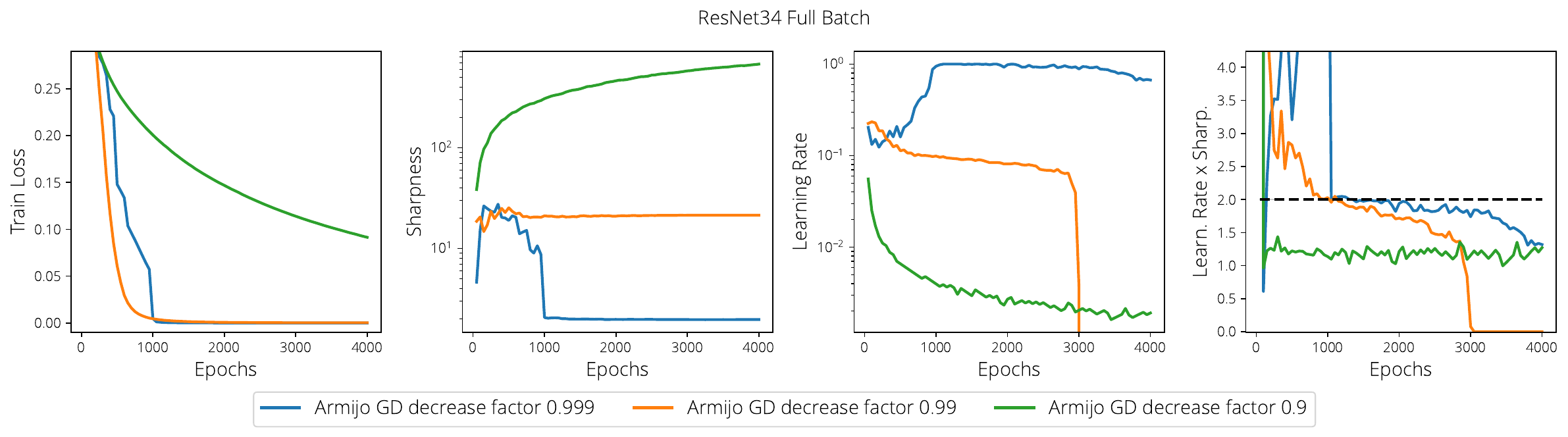}
  \caption{{\bfseries A more continuous adaptation of the learning rate ensures better convergence.}
  Armijo GD with tighter decreasing factors for a ResNet34, squared loss, on a
  subset of CIFAR10 (full batch). As the decreasing factor approaches 1., the
  Armijo linesearch avoids an ever-increasing sharpness and operates closer to EOS.
  \label{fig:armijo_continuous_decrease}
  }
\end{figure}

In Fig.~\ref{fig:armijo_continuous_decrease}, we consider choosing $c=0$, and
varying the decreasing factor used to guess the stepsize. In this experiment,
rather than guessing the stepsize at 1. at each iteration, we guess the stepsize
as $1.5$ times the previously accepted stepsize. We observe that tightening the
decreasing factor prevents the ever-increasing sharpness, while preserving a
monotonic decrease. This experiment hints that, by staying as close as possible
to the edge of stability we can prevent the sharpening issues observed with a
usual implementation of the Armijo linesearch with a decreasing factor of $0.8$
while remaining monotonic. However, adjusting the stepsize so precisely with a
decreasing factor close to $1$ is impractical as it requires many evaluations
steps. On the other hand, we observed selecting stepsizes larger than the EOS as
done by e.g. Polyak (Fig.~\ref{fig:zoom_polyak}) may not lead to divergent
behaviors while avoiding sharpening issues exhibited by Armijo GD. In other
words, it appears that a \emph{lower bound} on the stepsize is key to ensure
efficient optimization. An upper-bound on the other hand can always be
implemented with a loose Armijo implementation (Fig.~\ref{fig:armijo_loose}).

\subsection{Zoom into Polyak's behavior}
\begin{figure}
    \centering
    \includegraphics[width=\linewidth]{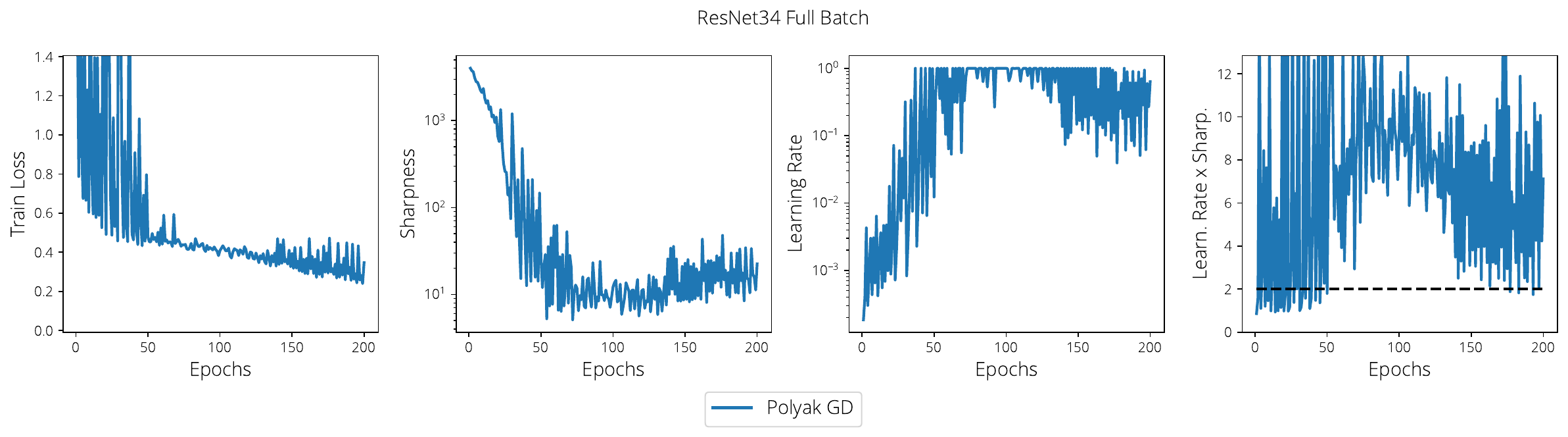}
    \caption{{\bfseries Warming up with aggresive stepsizes using Polyak
    stepsizes.} Polyak GD with maximal stepsize of 1. on a ResNet34 for the
    first 200 epochs, squared loss, subset of CIFAR10 (full batch). We observe
    that at the start, the learning rate remains small and progressively
    augments as in a warm-up scheme until the sharpness has sufficiently
    decreased for the optimizer to take large stepsizes.}
    \label{fig:zoom_polyak}
\end{figure}

In Fig.~\ref{fig:cifar_mlp_full_batch_lr_sharp} we plotted several metrics every
50 epochs. In Fig.~\ref{fig:zoom_polyak}, we show the train loss and the
learning rate at every epoch (that is every step as we are in full batch) for
the first 200 epochs. We observe that the learning rate is not just constant to
1 but oscillates between 0 and 1 except at the start where it stays at a low
value. Polyak starts taking large stepsizes (between 0 and 1) only once the
sharpness has significantly decreased.

\begin{figure}
    \centering
    \includegraphics[width=\linewidth]{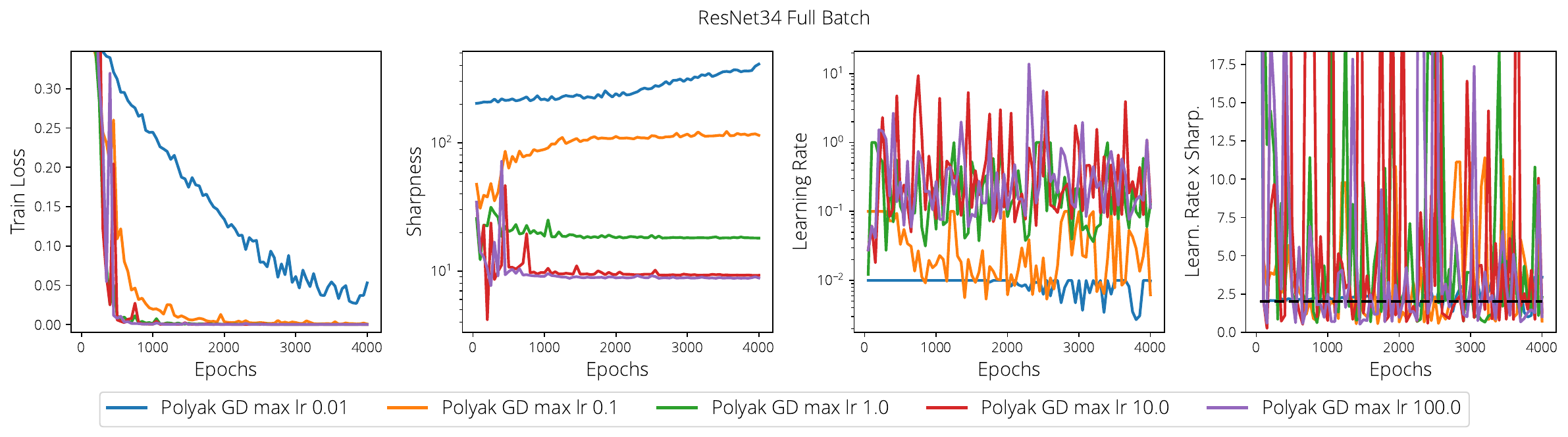}
    \caption{{\bfseries Polyak stepsizes are stable even without reaching maximal stepsize.}
    Varying the maximal stepsize of Polyak on a ResNet34, squared loss, subset of
    CIFAR10 (full batch). Even for large maximal stepsizes Polyak converges
    efficiently, though it is less stable at the start. Small maximum stepsizes
    lead to a similar behavior as constant stepsizes.}
    \label{fig:polyak_max_stepsizes}
\end{figure}

We also present on Fig.~\ref{fig:polyak_max_stepsizes}, how the maximal stepsize
for the Polyak method affects the increasing sharpness. Unsurprisingly by
setting a small maximal stepsize we retrieve a comportment akin to the one
displayed by constant stepsize SGD. On the other hand, in this experiment,
fixing a maximal stepsize even relatively large does not lead to divergence,
though the algorithm is less stable at the start.

\clearpage

\clearpage

\end{document}